# Convolutional capsule network for classification of breast cancer histology images

Tomas Iesmantas[1] and Robertas Alzbutas[1]

[1] Kaunas University of Technology, K. Donelaičio g. 73, Kaunas 44249
`tomas.iesmantas@ktu.lt`

**Abstract.** Automatization of the diagnosis of any kind of disease is of great importance and its gaining speed as more and more deep learning solutions are applied to different problems. One of such computer-aided systems could be a decision support tool able to accurately differentiate between different types of breast cancer histological images – normal tissue or carcinoma (benign, in situ or invasive). In this paper authors present a deep learning solution, based on convolutional capsule network, for classification of four types of images of breast tissue biopsy when hematoxylin and eosin staining is applied. The cross-validation accuracy, averaged over four classes, was achieved to be 87 % with equally high sensitivity.

## 1    Introduction

Breast cancer is one of the most frequent types diagnosed for women – it accounts for 30 % of all new cancer diagnoses in women [1]. However, it is a multifaceted disease with varying biological as well as clinical behaviors [2]. This heterogeneity resulted to an endeavor to classify this cancer into meaningful classes [3]. One may consider histological types, which refers to the growth patterns of the tumors, or molecular subtypes.
Histological grading is particularly important, because if the initial check-up for breast cancer (e.g. by palpation, mammography, ultrasound) is positive the breast tissue biopsies enables histological assessment of the severity of the cancer. However, histological analysis requires experience and extensive knowledge of the cytologist. Therefore, computer-aided decision systems would be of great help in detecting abnormalities and assessing their severity.

## 2    Related work

Advances of past decade in the deep learning techniques as well as computing power enabled systems for automatic classification of images: whether it is classification of many images found on internet of thousands general categories (see ImageNet competitions), to dermatologist-level skin cancer classification [4], to animal recognition in their habitats [5].

An attempt to apply deep learning techniques for breast cancer histological images has already been made – convolutional neural networks proved to be of great use in this task [6] allowing to achieve accuracies of 77.8 % for four class (normal, benign, *in situ* and invasive) and 83.3 % for carcinoma vs. non-carcinoma classification task. The accuracies achieved by convolutional neural network are truly high, considering that it requires no elaborate feature extraction methods before training the classifier – an advantage for which deep learning algorithms are often prized.

There are several other important examples of breast histological image analysis. Kowat et al. [10] used K-means, fuzzy C-means, competitive learning neural networks and Gaussian mixture models for nuclei segmentation and the results of this analysis were used in a medical decision support system for breast cancer diagnosis, where the cases were classified as benign or malignant (similar works were done by Filipczuk et al. [12] and George et al. [12]). Brooks et al. [13] considered a problem of classifying 361 images as normal, in situ and benign by support vector machines and achieved ~ 93 % accuracies for all classes. Zhang et al. [14] ensembles of SVM and neural networks to achieve 97 % classification accuracy for a 3-class (normal, benign and in situ) problem.

Above references are great examples of what machine learning/deep learning can achieve. In this paper a 4-class problem is considered: normal, benign, in situ and invasive types of histological images. In Materials and methods section, we briefly discuss the data and preprocessing steps together with more extensive presentation of Convolutional capsule networks (CapsNet) – a new type of networks [7].

## 3    Dataset and preprocessing

The dataset[1] is composed of hematoxylin and eosin (H&E) stained breast histology microscopy images. In total 400 images (in equal class proportions) was used. All images were of equal dimensions (2048 x 1536), with 0.42 μm x 0.42 μm pixel size. Each image is labelled with one of four classes: i) normal tissue, ii) benign lesion, iii) in situ carcinoma and iv) invasive carcinoma.

The images warried in the shading of the coloring probably due to slightly varying conditions and protocols of staining (see Fig. 1). Therefore, color transfer by Reinhard's method [9] was performed. In addition, to increase the number of trainable samples, three rotations of images were used: by 0 degrees (i.e. no rotation), by 90 and by 180 degrees. After an image rotation, 100 random patches of size 256 x 256 were cut. Hence, 300 patches were extracted from one image (3 rotations x 100 random patches).

---

[1] Additional test dataset was provided latter after the paper submission deadline.

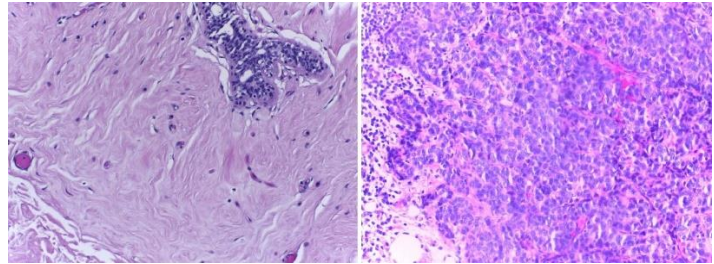

Fig. 1 Examples of different stain shades.

Because patches were generated randomly, no knowledge about the degree of overlap is retained. It is not clear whether such random cutting results to better performance. This aspect was not investigated any further.

## 4    Capsule Networks

Convolutional neural networks (CNN) suffer from several conceptual drawbacks: (1) max-pooling operation throws away information about the position of some entity that the network tries to recognize and (2) convolutional neural networks do not take into account many spatial relations between simpler objects. On the other hand, CNNs with max-pooling layers resulted to the rapid development of deep learning field. So, it was probably a matter of time till the method with CNN capabilities and without its disadvantages was developed - capsule network with dynamic routing [7]. The concept of capsules is not anything new, because G. E. Hinton, major figure in deep learning field, has been thinking about it for a while (see for example [8], although the idea goes back several decades ago, according to G. E. Hinton himself). It just never worked before, up until dynamic routing algorithm was proposed [7]. In what follows, the concept of Convolutional capsule network (CapsNet) will be presented in more details.

First of all, a capsule is a group of neurons whose outputs are interpreted as various properties of the same object. Each capsule has two ingredients: a pose matrix, and an activation probability. These are like activities of a standard neural network. The length of the output vector of a capsule can be interpreted as the probability that the entity represented by the capsule is present in the current input. There can be several layers of capsules. In our architecture, we used a layer of primary capsules (reshaped and squashed output of the last convolutional layer) and a layer of CancerCaps (i.e. capsules representing 4 types of images: normal/noncancerous, benign, in-situ and invasive).

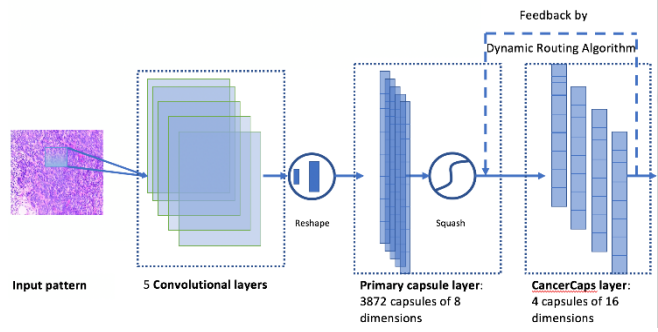

Fig. 2 Architecture of the Convolutional capsule network used to classify breast cancer histological images.

**Table 1** Considered Convolutional capsule network architecture.

|  | Layer type | Maps and neurons/capsules | Filter size/Strides<br>or<br>Capsule dimensions |
|---|---|---|---|
| 0 | Input | 3M x 512N x 512 N | 1 x 1 |
| 1 | Convolutional | 64M x 255N x 255 N | 4 x 4/2 |
| 2 | Convolutional | 128M x 126N x 126 N | 4 x 4/2 |
| 3 | Convolutional | 256M x 61N x 61 N | 6 x 6/2 |
| 4 | Convolutional | 256M x 28N x 28 N | 6 x 6/2 |
| 5 | Convolutional | 256M x 11N x 11 N | 8 x 8/2 |
| 6 | Primary capsule layer | 3872 C | 8 |
| 7 | CancerCaps layer | 4 C | 16 |

Before the layer of primary capsules, one can have as many convolutional layers as it fits. Only, the max-pool layers are missing; instead, to reduce the dimensionality, one used convolution with strides larger than 1 (if the stride is 2, then dimension are reduced by the factor of 2, etc.). The output of CancerCaps are used to make the decision about the class of the input image. An entire architecture of the network used in this work is presented in Fig. 2 and **Table** 1 contains information about the dimensions. The total number of trainable parameters was 9850816.

Each capsule in primary capsule layer is connected to every other capsule in CancerCaps layer. However, an algorithm, called routing-by-agreement, enables better learning as compared to the max-pooling routing. Routing-by-agreement is sort of a feedback algorithm which increases the contribution of those capsules which agree most with the parent output. Thus, even more strengthening its contribution.

The above-mentioned squashing function is a multidimensional alternative to the one-dimensional activation functions in regular neural networks (e.g. hyperbolic tangent, etc.) and is calculated as follows:

$$\boldsymbol{v}_j = \frac{\left\|\boldsymbol{s}_j\right\|^2}{1 + \left\|\boldsymbol{s}_j\right\|^2} \frac{\boldsymbol{s}_j}{\left\|\boldsymbol{s}_j\right\|},$$

where $\boldsymbol{v}_j$ is the vector output of capsule $j$ and $\boldsymbol{s}_j$ is its total input.

Another novelty introduced together with capsule networks was the use of margin-loss. For each cancer capsule, $k$ the incurred loss is as follows:

$$L_k = T_k \ \max(0, m^+ - \|\boldsymbol{v}_k\|)^2 + \lambda(1 - T_k) \ \max(0, \|\boldsymbol{v}_k\| - m^-)^2,$$

where $T_k = 1$ if and only if an image of class k is present and $m^+ = 0.9$ and $m^- = 0.1$. We use $\lambda = 0.5$.

## 5 Results

5-fold cross-validation was used with 25 % on whole images leaving for testing and the rest 75 % were used for network training. Adam optimizer [16] was used with parameter 0.0001 to train the entire network.

### 5.1 Image-wise classification

Image patches, due to the significantly smaller sizes than original images, were not all equally informative. Consider example in Fig. 3. The small patch in Fig. 3 (inside the black box) contains no information whether the entire image is taken from invasive carcinoma tissue or not. In other words, information contained in one large image is dispersed over the larger number of patches, some of which may not be of any value at all.

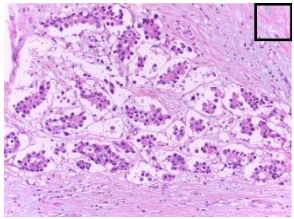

Fig. 3 Original image of invasive carcinoma and its patch (bounded by a black box)

Such information dispersion results to the noisy learning (i.e. loss function is noisy) and it is possible to quickly over-train the network with those uninformative patches. To avoid this the training was stopped when loss-function (computed on training samples) was less than 0.1. Only image-wise prediction was considered, i.e. accuracies were analyzed only for entire images and not for separate patches. The majority voting was used to decide on the label of the image.

Cross-validation procedure results were as in **Table** 2. The overall accuracy, as obtained from cross validation is 87%.

**Table 2** The confusion matrix for the cross-validation (mean values, %).

| True vs. Predicted | Benign | In situ | Inva-sive | Normal |
|---|---|---|---|---|
| Benign | 87 | 6 | 4 | 6 |
| In situ | 6 | 84 | 5 | 3 |
| Invasive | 5 | 5 | 88 | 1 |
| Normal | 2 | 5 | 4 | 90 |

The results on the test set of the competition was 72 %. However, some errors were made during the network training phase and therefore the results on the test set does not correspond to the properly trained network. At the time of submission of the paper competition organizers did not released the labels of the test set and the true results are not known (cross validation above corresponds to the correctly trained network).

Even though the cross-validation accuracies are high, it is clear that the network has difficulty to differentiate between Benign and Normal tissues. In addition, invasive type can be mixed up with benign and this type of mistakes can have severe consequences, as the invasive type of breast cancer requires immediate treatment.

### 5.2 Feature visualization

It is difficult to give any meaning to the different layers of the network and therefore it is not possible to understand clearly what gives the network the ability to discriminate between different classes. However, as exemplified in the Fig. 4 and **Error! Reference source not found.**, at least first convolutional layers try to recognize different parts of the histological image – nuclei, cytoplasm, and other objects. Going deeper into a network, the interpretability is lost due to the complexity of the network and calculations that it performs.

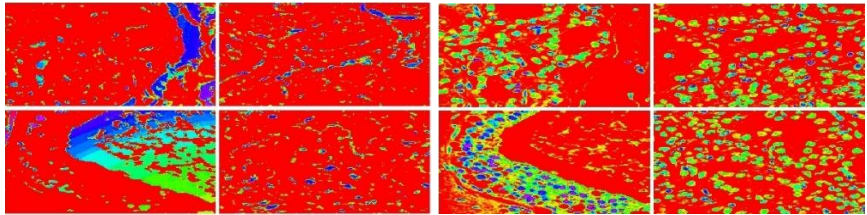

Fig. 4 Example on the left of first convolutional layer features where other than nuclei and cytoplasm areas are enhanced. Example on the right of first convolutional layer features, where nuclei are enhanced

It is also interesting to look at the visualization of the output of CancerCaps layer (i.e. layer consisting of 4 cancer capsules of 16 dimensions). For this purpose, t-SNE method [15], a parametric embedding technique for dimensionality reduction, was applied.

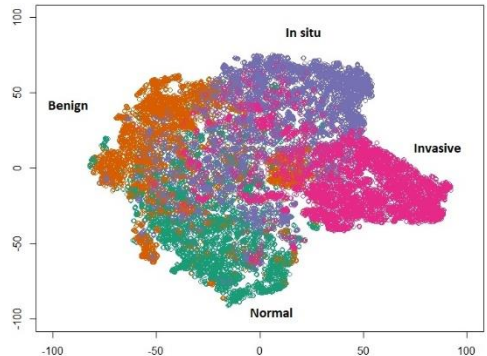

Fig. 5 t-SNE visualization of CancerCaps layer features.

In Fig. 5 a visualization of how different four classes are. All image classes overlap significantly. But this is expected because, as was noted previously, an entire histology image was divided into much smaller patches, many of which carried no information about the specific class or that information was misleading.

## 6    Conclusions and further discussion

A convolutional capsule network was presented to solve the classification task of breast cancer histological images. The cross-validation accuracy was 87 % for the benign carcinoma tissue images, 84 % for the *in situ* carcinoma, 88 % for invasive type and 89 % for normal tissue images. As of now, testing phase results remain unknown and will be added shortly.

It is unknown whether different variations of network architecture would've resulted to similar or better results. In the future, more in depth analysis will be performed to optimize the architecture: number of convolutional layers, dimensions of capsules in primary and CancerCaps layers. Also, no regularization was considered, although decoding part of autoencoder after capsule layers was suggested to have a positive impact on the learning and generalization. However, cross-validation results are very promising, hinting that capsule networks are of equal or even better capabilities as compared to the classical convolutional neural networks.

It is probably safe to speculate, that a computer-aided decision system, which would help to diagnose breast cancer faster and more accurately, can be envisaged in a near future.

## Acknowledgement

Tomas Iesmantas was supported by the Postdoctoral fellowship (supervisor Robertas Alzbutas) grant kindly provided by the Kaunas University of Technology and the Faculty of Mathematics and Natural Sciences.